\documentclass[a4paper]{article}
\pdfoutput=1
\usepackage{INTERSPEECH2021}
\usepackage{url}
\usepackage{arydshln}

\title{Usefulness of Emotional Prosody in Neural Machine Translation}
\name{Charles Brazier \hspace{0.5cm} Jean-Luc Rouas}
\address{Univ. Bordeaux, CNRS, Bordeaux INP, LaBRI, UMR 5800, F-33400 Talence, France}
\email{charles.brazier@u-bordeaux.fr \hspace{0.5cm} jean-luc.rouas@labri.fr}

\usepackage{color}

\begin{document}

\maketitle
\begin{abstract}
Neural Machine Translation (NMT) is the task of translating a text from one language to another with the use of a trained neural network. Several existing works aim at incorporating external information into NMT models to improve or control predicted translations (e.g. sentiment, politeness, gender). In this work, we propose to improve translation quality by adding another external source of information: the automatically recognized emotion in the voice. This work is motivated by the assumption that each emotion is associated with a specific lexicon that can overlap between emotions. Our proposed method follows a two-stage procedure. At first, we select a state-of-the-art Speech Emotion Recognition (SER) model to predict dimensional emotion values from all input audio in the dataset. Then, we use these predicted emotions as source tokens added at the beginning of input texts to train our NMT model. We show that integrating emotion information, especially arousal, into NMT systems leads to better translations.
\end{abstract}
\noindent\textbf{Index Terms}: Neural Machine Translation, Speech Emotion Recognition, Neural Networks.

\section{Introduction}
\label{sec:introduction}

Machine Translation (MT) is the task of translating a text from one language to another. Existing methods are originally based on rule-based or statistical approaches. More recently, end-to-end neural networks have demonstrated their efficiency in the task \cite{sutskever2014sequence, bahdanau2014neural}, outperforming earlier methods \cite{kocmi2023findings}, and leading to the emergence of the Neural Machine Translation (NMT) task.

Transformer-based architectures \cite{vaswani2017attention} are widely used in NMT, leading to state-of-the-art performances for several language pairs. Transformers are sequence-to-sequence models composed of an encoder that reads the input sequence and generates a representation of it, a decoder that generates the output sequence based on the input representation, and self-attention mechanisms that enable the use of the entire input sequence for each output prediction.

Several works in the domain focused on the control of the translation given by existing NMT models. In fact, most NMT models aim to generate a single correct translation of a given sentence. However, the translation can be dependent on specific information that may be missing in the input sentence. For example, translations are dependent on the speaker's gender that may not be specified in the source language, or even some words can be translated differently depending on the context. For this purpose, several models propose solutions to improve translation quality by controlling sentiment \cite{si2019sentiment}, politeness \cite{sennrich2019controlling}, gender \cite{vanmassenhove2018getting, gaido2023how}, style of translators \cite{wang2021towards}, or output language in the case of a multilingual setup \cite{johnson2017google}.

In this work, we propose to improve translation quality by adding another external source of information: the automatically recognized emotion in the voice. This work is motivated by the assumption that each emotion is associated with a specific lexicon that can overlap between emotions. One possible approach to estimate the emotion of a specific sentence is to use affective word lists that classify individual words into emotion categories \cite{pennebaker2001linguistic} or rate them on emotion dimension scales for arousal, dominance, and valence \cite{warriner2013norms}. However, the emotion of words is dependent on the entire sentence context \cite{braun2020affective}. Additionally, word lists are not exhaustive and are specific to particular languages. In the following, we focus on the prosody of the pronounced sentences present in audio recordings to estimate emotion.

To this end, we first select a state-of-the-art Speech Emotion Recognition (SER) model to predict dimensional emotion values from audio recordings. Then, we use these predicted emotions as source tokens added at the beginning of input texts to train our NMT model. We show that integrating emotion information, especially arousal, into NMT systems leads to better translations.

In this paper, existing works adding external information to NMT models will be presented in Section~\ref{sec:Related_works}. The selected state-of-the-art SER model and our proposed emotion-aware NMT pipeline will be described in Section~\ref{sec:EmotionAware_NMT}. Experiments and results will be presented and discussed in Section~\ref{sec:Experiments}. Section~\ref{sec:Conclusion} will conclude the work.

\section{Related works}
\label{sec:Related_works}

Existing works aiming at incorporating external information into NMT models to improve or control translation quality are numerous. They generally involve a two-stage procedure. At first, the external information is translated into a special token with the help of manual or automatic annotations. Then, the special token is added into the NMT model to control translation.

In the case of sentiment, Si et al. \cite{si2019sentiment} use a trained sentiment classifier to label ambiguous words with positive or negative tags. Then, the authors incorporate the sentiment information into the source sentence either at the beginning or directly before the ambiguous word. They also try to use two different embedding vectors for each ambiguous word and select the correct embedding depending on the desired sentiment.

In the case of politeness, Sennrich et al. \cite{sennrich2019controlling} automatically annotate politeness (informal/polite) using linguistic rules and add the special token at the end of the source sentence.

In the case of gender, Vanmassenhove et al. \cite{vanmassenhove2018getting} extract gender information (male/female) from meta data present in their dataset and use it as additional token added at the beginning of the source sentence.

In the case of translation style, Wang et al. \cite{wang2021towards} use the data translated by 10 translators and incorporate the translator token at the beginning of input sentences, or add the embedded translator token to every token in the encoder. They also try to add the translator information directly into the decoder but this solution seems ineffective.

In the case of a multilingual scenario, Johnson et al. \cite{johnson2017google} add an artificial token at the beginning of input sentences to specify the desired target language.

Our approach aims at enhancing translation quality by adding emotion information into text sentences. Emotion is automatically extracted from audio data (recordings or synthesized text) with the help of a trained SER model and added to source sentences at the beginning as an additional emotion token.

\section{Emotion Aware Neural Machine Translation}
\label{sec:EmotionAware_NMT}

In this work, we propose to combine SER classifier and NMT models to improve translation quality.

To the best of our knowledge, the only work that combines an emotion classifier with an NMT model is presented by Troiano et al. \cite{troiano2020lost}. The authors introduce a method to preserve emotion during automatic translation by re-ranking the best translations generated by an NMT model, according to a trained emotion classifier. Due to the lack of comparable emotion classifiers for different languages, they adopt a back-translation setup that re-translates the best translation hypotheses and re-rank them to preserve the same emotion between the input sentence and its translated translations. The re-ranking is performed as a post-processing step and not as a fine-tuning procedure for the NMT model.

In contrast to \cite{troiano2020lost}, we aim at improving translation quality by using the output of a pre-trained emotion classifier in the NMT training step.

Our proposed method follows a two-stage procedure. At first, a trained Speech Emotion Recognition (SER) model predicts dimensional emotion values from all input audio in the dataset. Then, these predicted values are converted into discrete tokens and added at the beginning of input texts. Finally, we train various NMT models by varying the utilized emotion token and extraction method (whether it is predicted from original or synthesized recordings).

\subsection{Speech Emotion Recognition}
\label{subsec:SER}

Automatic Speech Emotion Recognition (SER) is the task of predicting emotion from speech recordings. SER models can predict discrete categories such as happy, sad, etc., or continuous emotional dimensions like arousal, valence, and dominance. The three emotional dimensions are described as follows:
\begin{itemize}
    \item Arousal refers to the level of stimulation or activation associated with an emotion. It ranges from low arousal (calm, relaxed) to high arousal (excited, agitated). Emotions like excitement, anxiety, and surprise typically have higher arousal levels, while calmness and relaxation have lower arousal levels.
    \item Valence represents the positive or negative nature of an emotion. Emotions can be categorized on a scale from positive (happy, joyful) to negative (sad, angry). Valence indicates whether an emotion is pleasant or unpleasant, with positive valence indicating positive emotions and negative valence indicating negative emotions.
    \item Dominance refers to the level of control or power associated with an emotion. It measures whether an emotion makes an individual feel dominant or submissive in a particular situation. Emotions with high dominance might include feelings of confidence, assertiveness, or control, while low dominance emotions might involve feelings of submission or helplessness.
\end{itemize}

These dimensions are often used in psychology and neuroscience to help understand and categorize emotions, providing a more nuanced view of the complex nature of human feelings and experiences.

For our experiments, we select a state-of-the-art transformer-based SER model proposed by Wagner et al. \cite{wagner2023dawn}. This model is based on the wav2vec2.0 \cite{baevski2020wav} model which learns powerful speech representations from unlabelled speech data. Wav2vec2.0 is composed of 7 Convolutional Neural Networks (CNN) layers of feature encoder and 12 transformer layers of encoder including multi-head self-attention modules and fully-connected layers. Wagner et al. add one average pooling layer, one hidden layer, and one output layer of length 3 at the end of the network to classify each raw waveform into three emotion dimensions corresponding to arousal, dominance, and valence. The three output values range from 0 to 1.

During training, the authors freeze the CNN layers and only fine-tune transformer layers with the added layers on top of the model. The fine-tuning step is performed on the MSP-Podcast \cite{lotfian2017building} dataset (train split) for training, and tested on MSP-Podcast (test-1 split), IEMOCAP \cite{busso2008iemocap}, and MOSI \cite{zadeh2016multimodal}. All three corpora contain audio recordings with corresponding dimensional emotion values that are normalized between 0 and 1. Model performances are evaluated with the Concordance Correlation Coefficient (CCC) \cite{atmaja2021evaluation}. The authors report a CCC of .744 for arousal, .655 for dominance, and .638 for valence, on the MSP-Podcast dataset.

\subsection{Neural Machine Translation}
\label{subsec:NMT}

Neural Machine Translation (NMT) is the task of translating a text from one language to another. As model for our experiments, we selected a transformer-based encoder-decoder architecture developed with the ESPnet toolkit \cite{watanabe2018espnet}. The model is composed of an encoder of 6 transformer layers, a decoder of 6 transformer layers, and 4 attention heads in each self-attention layers. The model is end-to-end meaning that it receives text as input and outputs text. All textual sentences are tokenized into sub-word sequences using a byte-pair encoding algorithm with 1000 units.

The model is trained on the Libri-trans \cite{kocabiyikoglu2018augmenting} dataset, a dataset that contains speech recordings and corresponding transcripts in English and French. The dataset is divided into train, dev, and test sets whose durations are respectively 230, 2, and 3.5 hours. In this work, we focus on text-to-text translation from English to French.

The model is evaluated with BiLingual Evaluation Understudy (BLEU) score, a metric that measures the similarity between generated translations and corresponding groundtruth. Its calculation includes a brevity penalty, penalizing generated translations shorter than their references, and a n-gram precision factor that considers various n-grams (including unigrams, bigrams, etc.) and counts the number of n-grams from the generated translation that also appear in the reference translation. The BLEU score is a value between 0 (poor translation) and 100 (perfect translation). Note that a translation with a BLEU score higher than 40 is generally considered as a high-quality translation. We report BLEU using SacreBLEU \cite{post2018call}.

\subsection{Emotion Aware Neural Machine Translation}
\label{subsec:EmotionAwareNMT}

To improve the translation quality of our NMT model, we propose to incorporate the emotion information extracted from input recordings into their corresponding input text sequences. The emotion information is represented as an additional token representing the polarity over a specific emotion dimension. We use the tokens $<$AroNeg$>$ and $<$AroPos$>$ for recordings that have an arousal score lower or higher than 0.5 respectively, as well as $<$DomNeg$>$ and $<$DomPos$>$ for dominance, and $<$ValNeg$>$ and $<$ValPos$>$ for valence.

For example, consider the following pair of English and French sentences:
\newline \centerline{
\texttt{\footnotesize I am quite foolish} $\rightarrow$ \texttt{\footnotesize Je suis toute sotte}}
In the case of valence, the pair will be modified to:
\newline \centerline{
\texttt{\footnotesize $<$ValNeg$>$}  \texttt{\footnotesize I am quite foolish} $\rightarrow$ \texttt{\footnotesize Je suis toute sotte}}

The proposed method offers several advantages. Firstly, it does not necessitate any changes in the model architecture nor in the set of translated sequences. Model performances with and without the additional token can be directly compared. Furthermore, the approach is simple yet effective, showcasing progress in various domains \cite{si2019sentiment, sennrich2019controlling, vanmassenhove2018getting, wang2021towards, johnson2017google}. It only involves the addition of an extra token at the beginning of each input sentence. Lastly, the method does not require any additional manual annotation, as the emotion is automatically extracted from the SER model.

\section{Experiments and results}
\label{sec:Experiments}

In this section, we conduct a set of experiments to compare NMT performances under different configurations. Firstly, we analyze the statistical distribution of values provided by the SER model on the Libri-trans dataset for the three emotion dimensions. Secondly, we train four different NMT models that are a baseline model, and three different models that use the arousal, dominance, and valence information respectively.

\subsection{Emotion recognition on Libri-trans}
\label{subsec:EmotionRecognition_Libritrans}

Our proposed method starts with the computation of emotional dimensions for each English sentence in the Libri-trans dataset. These values will be later used for the English-to-French text-to-text translation task.

\begin{figure}[t]
  \centering
  \includegraphics[width=\linewidth]{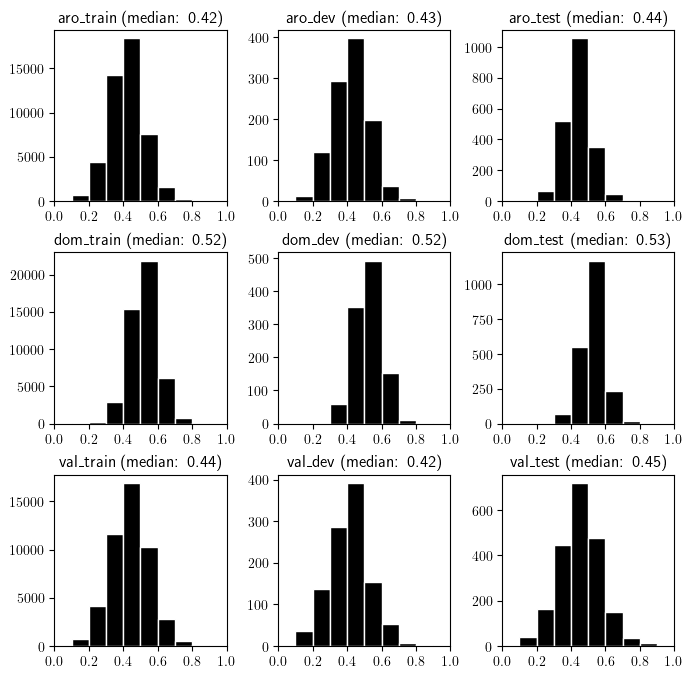}
  \caption{Statistical distribution of emotion values on original audio from the Libri-trans dataset.}
  \label{fig:distribution}
\end{figure}

Due to the modality mismatch between the text-to-text task and the required raw audio waveform as input for our SER model, we propose two different ways to compute emotion values. The first method involves using audio recordings that are already present in the Libri-trans dataset to evaluate emotion values. This represents the most accurate method to estimate emotion; however, it implies the use of additional data in the translation pipeline. Statistical distributions of emotion values on original audio recordings from the Libri-trans dataset are illustrated in Figure~\ref{fig:distribution}. We showcase statistical distributions for each train, dev, and test subset of the Libri-trans dataset. We remark that, for the three emotional dimensions and each subset, statistical distributions are balanced, medians are around 0.5 and values range from 0.1 to 0.9. Due to the acceptable balance of dimensional emotion scores provided by our selected SER, we consider the initial train, dev, and test splits as good candidates for the translation task conditioned with emotion information.

\begin{figure}[t]
  \centering
  \includegraphics[width=\linewidth]{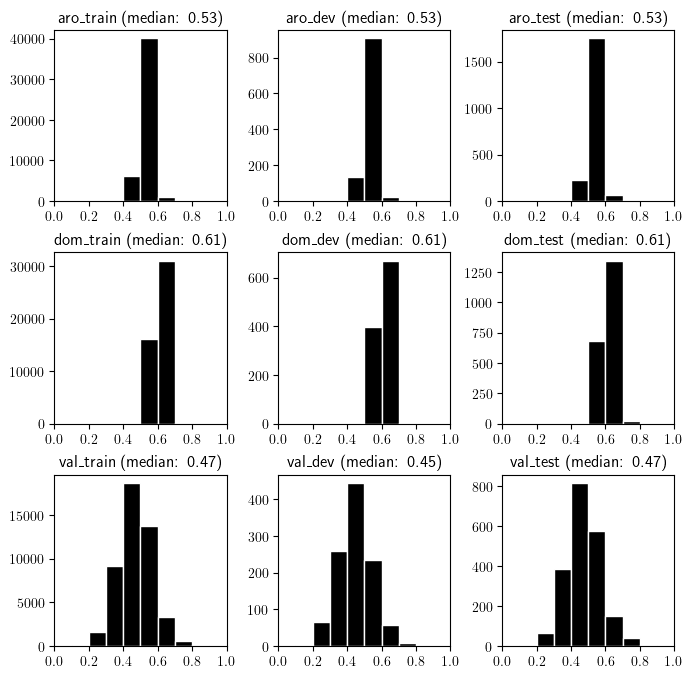}
  \caption{Statistical distribution of emotion values on synthesized audio from the Libri-trans dataset.}
  \label{fig:distribution_synth}
\end{figure}

The second method involves synthesizing each English text with the help of the Google Text-to-Speech API\footnote{\url{https://gtts.readthedocs.io/en/latest/}}. This method has the advantage of using only text data; however, the original emotion can be altered with a neutral rendering. Statistical distributions of emotion values on synthesized audio of all English text are illustrated in Figure~\ref{fig:distribution_synth}. We observe that arousal values are concentrated around 0.5, indicating that synthesizing audio files produces neutral arousal. Also, we remark that almost all dominance values are above 0.5. Thus, using synthesized audio to estimate dominance is inefficient. However, valence values are more distributed, reflecting the fact that valence is captured in the vocabulary used in each English sentence. As an additional observation, the medians of arousal and valence values across all Libri-trans subsets are close to 0.5, enabling an automatically balanced dataset for training.

\subsection{Neural Machine Translation on Libri-trans}
\label{subsec:NMT_Libritrans}

The computation of emotion values on English data from Libri-trans enables the assignment of a specific dimension polarity for each input sentence. This is reflected by the addition of an extra token at the beginning of each input text.

In the following experiments, we compare four different NMT models. The first model, \textit{baseline}, is trained without the use of an emotion token and serves as a reference to improve its translation score. The second, third, and fourth models, namely \textit{arousal}, \textit{dominance}, and \textit{valence}, are NMT models trained with the addition of a special token at the beginning of each English sentence representing the polarity (positive or negative) of their corresponding emotion dimension.

We also conduct two sets of experiments, one using tokens extracted from original audio recordings and the other using those extracted from synthesized audio.

All models have the exact same structure. They are trained with a label-smoothing loss, a Noam optimizer, a dropout rate of 0.1, an inverse square root scheduler with 8000 warm-up steps, and a batch size of 96. Each transformer model is trained during 250 epochs and the 5 best MT models are averaged to create the best model. Each training session takes approximately 15 hours on a single NVIDIA A100 GPU.

Table~\ref{tab:BLEUscores} showcases the results of our different experiments. In this table, we report BLEU scores of 8 different experiments. It gathers the evaluation of our four proposed models, evaluated on the development and test sets of Libri-trans, with the emotion token computed either from the synthesized text or the audio recordings directly.

\begin{table}[t]
  \caption{BLEU scores of our four proposed models on dev and test sets of Libri-trans. The emotion token is computed either from the audio recordings of Libri-trans (Recor) or the synthesized text (Synth).}
  \label{tab:BLEUscores}
  \centering
  \begin{tabular}{lcccc}
    \toprule
    \textbf{Model} & \textbf{baseline} & \textbf{arousal} & \textbf{dominance} & \textbf{valence}\\
    \midrule
    Recor dev  & 20.1 & 20.2 & 20.5 & 20.6\\
    Recor test & 18.2 & \textbf{18.6} & 18.3 & 18.2\\
    \hdashline
    Synth dev  & 20.1 & 20.2 & 20.1 & 20.1\\
    Synth test & 18.2 & \textbf{18.3} & 18.0 & 17.9\\
    \bottomrule
  \end{tabular}
\end{table}

Regarding the baseline model, we emphasize that it does not include any additional token, and thus, BLEU scores are insensitive to the type of modality used to extract the emotion token and are respectively 20.1 and 18.2 on the dev and test sets. 

In comparison with the baseline performances, we observe that the addition of the emotion token computed from real speech leads to a noticeable improvement for arousal, with a BLEU score of 18.6 on the test set of Libri-trans, and there is no degradation in BLEU score for dominance and valence, with BLEU scores of 18.3 and 18.2 respectively. These results show that the arousal information seems to be a promising factor to enhance translation quality. The best performances are achieved when using the arousal information, which has been automatically estimated from real speech, resulting in a 0.4-point improvement in the BLEU score.

When the emotion token is computed from synthesized text, we observe a slight improvement in the BLEU score for arousal. However, for dominance, we notice a slight degradation in the BLEU score. As mentioned earlier, dominance value estimations on synthesized audio almost always provide a value higher than 0.5. Thus, we anticipated a BLEU score lower or equal to the baseline, as no additional information is introduced to the NMT model. Finally, the highest deterioration in BLEU score is observed for valence.

Using emotion tokens extracted from real speech always leads to better BLEU scores than when using tokens from synthesized speech. We observed that, with synthesized audio, emotion value estimations become neutral in the case of arousal and dominance, and translation quality is degraded. Therefore, it appears crucial to use real audio recordings to estimate emotion and improve translation quality.

In addition, we note that all BLEU scores presented in Table~\ref{tab:BLEUscores} are consistently low, indicating poor translation quality for all proposed models, including the baseline. This is due to the nature of the data within the Libri-trans dataset, which comprises English utterances extracted from books. Audios are recorded from readers, and the vocabulary in the texts differs significantly from the language commonly spoken in real-world interactions.

\section{Conclusion}
\label{sec:Conclusion}

We proposed an innovative method that combines SER with NMT models to improve translation quality. Our method is fully automatic and does not require any manual annotation. The proposed method is simple, effective, and enables the direct comparison between models. The best performances were achieved by adding the arousal token extracted from real speech at the start of each input sentence.

This study only presents the first experiment on combining SER and NMT models to improve translation. Additional experiments can be conducted, such as using the three emotional dimensions to condition each source sentence, switching languages (translating from French to English), testing the method on other multilingual datasets including MuST-C, or visualizing the embedding of the NMT model to get a better understanding on how vocabulary related to a specific emotion is clustered. Moreover, different methods to include the emotion information can be tested, such as adding the embedding of the emotion token to all token embeddings.

In this work, we incorporate real speech recordings as an additional modality to estimate emotion. Introducing the audio modality into the translation task also enables to perform speech-to-text translation, known as Speech Translation. Consequently, exploring the conditioning of Speech Translation models with emotion as additional information would be valuable.

\section{Acknowledgements}
\label{sec:Acknowledgements}

The research presented in this paper is conducted as part of the project FVLLMONTI, which has received funding from the European Union’s Horizon 2020 Research and Innovation action under grant agreement No 101016776.

\bibliographystyle{IEEEtran}
\bibliography{mybib}
\end{document}